\documentclass[a4paper,twoside]{article}

\usepackage{epsfig}
\usepackage{subcaption}
\usepackage{calc}
\usepackage{amssymb}
\usepackage{amstext}
\usepackage{amsmath}
\usepackage{amsthm}
\usepackage{multicol}
\usepackage{pslatex}
\usepackage{apalike}
\usepackage[bottom]{footmisc}

\usepackage[hidelinks]{hyperref}
\usepackage{booktabs}
\usepackage{array}
\usepackage{cite}
\usepackage{siunitx}
\usepackage{stfloats}
\usepackage{xcolor}
\usepackage[british]{babel}

\usepackage{SCITEPRESS}     % Please add other packages that you may need BEFORE the SCITEPRESS.sty package.

\begin{document}

\title{Towards Audit Requirements for AI-based Systems in Mobility Applications}
\author{\authorname{
		Devi Padmavathi Alagarswamy\sup{1},
		Christian Berghoff\sup{2},
		Vasilios Danos\sup{3},
		Fabian Langer\sup{3},
		Thora Markert\sup{3}\sup{*},
		Georg Schneider\sup{4},
		Arndt von Twickel\sup{2}\sup{+} and
		Fabian Woitschek\sup{4}\sup{+}\sup{*}
}
\affiliation{\sup{1}ZF Friedrichshafen AG, System House Autonomous Mobility Systems, Friedrichshafen, Germany}
\affiliation{\sup{2}German Federal Office for Information Security (BSI), Bonn, Germany}
\affiliation{\sup{3}TÜV Informationstechnik GmbH, IT Security Hardware Evaluation, Essen, Germany}
\affiliation{\sup{4}ZF Friedrichshafen AG, Artificial Intelligence Lab, Saarbr\"ucken, Germany}
\affiliation{\sup{+}Contact: fabian.woitschek@zf.com, arndt.twickel@bsi.bund.de}
\affiliation{\sup{*}Main Contribution}
}


\abstract{
	Various mobility applications like advanced driver assistance systems increasingly utilize artificial intelligence (AI) based functionalities. 
	Typically, deep neural networks (DNNs) are used as these provide the best performance on the challenging perception, prediction or planning tasks that occur in real driving environments. 
	However, current regulations like UNECE R 155 or ISO 26262 do not consider AI-related aspects and are only applied to traditional algorithm-based systems. 
	The non-existence of AI-specific standards or norms prevents the practical application and can harm the trust level of users. 
	Hence, it is important to extend existing standardization for security and safety to consider AI-specific challenges and requirements. 
	To take a step towards a suitable regulation we propose 50 technical requirements or best practices that extend existing regulations and address the concrete needs for DNN-based systems. 
	We show the applicability, usefulness and meaningfulness of the proposed requirements by performing an exemplary audit of a DNN-based traffic sign recognition system using three of the proposed requirements.
}

\onecolumn \maketitle \normalsize \setcounter{footnote}{0} \vfill

\section{\uppercase{Introduction}}
\label{sec:introduction}

Artificial intelligence (AI) -based systems are increasingly used as part of mobility applications like autonomous driving (AD) or advanced driver assistance systems (ADAS).
Especially, deep neural networks (DNNs) achieve an impressive performance on most tasks and are the most promising solution to achieve higher levels of automated driving.
At the same time, different manufacturers already use DNN-based solutions as part of ADASs with partial automation (SAE L2 \cite{SAEJ3016}) \cite{tesla} that are operating on public roads or for highly automated shuttles (SAE L4) \cite{waymo} operating in limited public areas.
However, current DNNs introduce new and specific vulnerabilities into the systems which can impact the performance and trustworthiness of AD/ADAS systems negatively.
This requires a detailed analysis of existing vulnerabilities and potential mitigation strategies.
To still enable the usage of such DNN-based solutions for high-risk applications, like highly or fully automated driving (SAE L4/L5), clear guidelines and regulations are required.
This assures that systems with a high degree of autonomy are trustworthy with respect to use case relevant aspects like safety, security, robustness or explainability and include mitigation strategies to known vulnerabilities.

However, currently no homologation regulations or standards exist that are tailored towards the use of AI-based systems in mobility applications and include AI-specific vulnerabilities \cite{Radlak20}.
There are no uniformly acknowledged principles and practices that the development, testing or deployment of AI-based systems must fulfill. 
This limits the future deployment of AI-based systems to low-risk applications.
Furthermore, it represents a major challenge for industry, auditors and regulators and potentially leads to a lower level of user trust.

To provide guidance for future regulations, in this work we explore how auditing guidelines can ensure the security and safety of AI-based systems in high-risk applications.
Thereby, we focus on the application of AI-based systems for mobility applications like functionalities for AD/ADAS.
Hence, we base our work on existing standards for road vehicles, like the functional safety standard ISO 26262 \cite{ISO26262}, which are relevant for mobility applications, and provide the following contributions:

\begin{itemize}
	\item We present an overview of existing and developing standards relevant for auditing AI-based systems in mobility applications.
	\item We introduce a list of generic requirements which focus on specific needs arising when auditing AI-based systems.
	\item We compare different use cases for AI-based systems in mobility applications to select the most suitable use case which is used for testing and refining the introduced audit requirements in practice.
	\item We demonstrate the applicability of the most relevant audit requirements for the selected use case.
\end{itemize}

\section{\uppercase{Related Work}}
\label{sec:related-work}

During an audit the compliance to industry standards mandated by regulators is evaluated.
However, traditional standards for systems in mobility applications do not contain specific guidelines in case AI-based systems are utilized instead of traditional algorithm-based systems.
This includes both safety standards like \cite{ISO26262}, \cite{ISO21448} or \cite{ANSI4600} and security standards like \cite{ISO21434} or \cite{UNECER155}.

Since currently no standards for the auditing of AI-based systems exist, there are approaches to develop an appropriate standardization. 
Best known here is the \cite{AIAct} which tries to lay down uniform regulations for AI-based systems. 
It presents a horizontal regulatory approach with necessary requirements to address different risks and challenges when AI is used, without focusing on the needs for specific application areas. 
Similarly, \cite{ISO24028} focuses on the trustworthiness of AI-based systems without considering a concrete application domain. 
It surveys different generic threats and risks and also covers existing mitigation strategies. 
Additionally, \cite{ISO24029} provides background information on existing methods to assess the robustness of generic DNNs.

In addition to the already published drafts for standardization of AI-based systems mentioned above, there are also ongoing standardization activities.
This includes \cite{ISO5469}, which covers aspects of functional safety specific for AI-based systems.
In addition, \cite{ISO4213} focuses on methods to assess the performance of ML-based classification systems.
Furthermore, there are standardization activities on a national level which include the \cite{DINAI}.
Here, requirements and challenges as well as standardization needs for seven topics around AI are discussed.

In contrast to these horizontal regulatory approaches, there are also vertical regulatory approaches. 
These aim to develop standards for concrete application areas and multiple standards are in development for the usage of AI in (high-risk) mobility applications. 
Here, \cite{ISO5083} gives guidance of the steps for developing and validating safe AD/ADAS systems.
It covers the SAE levels L3/L4 and the impact of using AI-based systems as part of larger AD/ADAS systems. 
Additionally, \cite{ISO8800} focuses specifically on the interaction between safety and AI. 
It defines risk factors for vulnerabilities in the behavior of AI within mobility applications.

Finally, there are only very few publications available that focus on the auditing of AI systems in practice.
Here, \cite{Raji20} introduces a framework for auditing AI-based systems throughout the internal development lifecycle.
This results in a series of documents which form an overall audit report that can be used by auditors for a formal audit.

After discussing available and developing standards for AI-based systems in general and specific to mobility applications, we now shortly present an overview of related publications which focus on specific vulnerabilities or challenges that exist for AI-based systems.
In \cite{Li22} the authors present a summary of important aspects for trustworthy systems which they deem necessary to be audited. 
\cite{Berghoff20} focuses on discussing known vulnerabilities of current AI-based systems and how they compare with traditional algorithm-based systems. 
Similarly, \cite{Mohseni20} first discusses current vulnerabilities specific to AI-based systems and then present possible mitigation strategies.

Concretely, there are multiple new security challenges for AI-based systems. 
These include model extraction attacks \cite{Papernot17, Orekondy19}, where an adversary attempts to copy the functionality of a victim AI model. 
%Carlini19
Next, evasion attacks, also known as adversarial attacks \cite{Szegedy14, Madry18}, are carefully perturbed input samples (adversarial examples) that change the prediction of AI-based systems according to the will of an adversary. 
This imposes a threat on the integrity of the system.
Lastly, data poisoning attacks \cite{Goldblum20, Schwarzschild21} describe the injection of poisoned data samples in the training dataset of an AI-based system. 
This degrades the behavior of the resulting system depending on the specific goals of the adversary.

In addition to security related AI-specific vulnerabilities there are also challenges regarding the robustness against natural perturbations \cite{Hendrycks19, Geirhos20}.
Concretely, out-of-domain data describes the presence of data samples that deviate from the exact training distribution used during training of an AI-based system.
This effect occurs naturally when systems are deployed in the real-world outside of a completely supervised environment and presents a challenge on the generalization of such systems.

Also, the existing black-box character of AI-based systems, combined with the complexity and number of parameters of DNNs, complicates the possibility to explain the behavior of a system.
It is largely unclear how a system arrives at its predictions and which features of a data sample are most important for a concrete prediction.
Therefore, the need arises for methods that can explain the decision or general behavior of a system that is learned from data.
These methods are published under the term of explainable artificial intelligence \cite{Gilpin18, Guidotti18}.

\section{\uppercase{Generic Audit Requirements}}
\label{sec:requirements}

In the following, generic requirements are derived based on a detailed analysis of established security and safety standards. 
As stated in \autoref{sec:related-work} currently there are no existing AI-specific certification standards, norms or regulations for systems in mobility applications. 
The available standards and norms are designed for traditional algorithm-based automotive systems.
We extract their AI-relevant aspects and complement them with AI-specific formulations.

\subsection{Requirements Elicitation}

Due to the special characteristics of AI components (e.g. high data complexity, non-linearity or lack of interpretability), some of the existing standards may not apply to such components or have to be adjusted to also ensure the safety and security of AI-based systems.
Hence, the generic requirements are formulated to address the technical aspects performance, robustness, explainability, external monitoring and the documentation of the entire mobility system and its AI subsystems. 
Furthermore, we consider requirements along the entire lifecycle of such systems. 
Fairness and privacy are out of scope for this work and should be addressed in future research. 
In total we formulate 50 generic requirements, which are available in the project report at \cite{P538}.
In the following, we present our general approach to derive the requirements and discuss three exemplary requirements which can be applied to most AI-based systems and are highly relevant.

The \cite{ISO26262} introduces the automotive safety level integrity (ASIL), which is a risk level based classification of recommendations for automotive systems. 
In this classification scheme the system's risk level is categorized in four risk levels, through the possible exposure to hazards, the controllability of a hazard and the severity of possible injuries to the driver and passengers stemming from the hazard. 
The four ASIL levels range from “ASIL A” associated with the lowest degree of risk to “ASIL D” associated with the highest degree of risk. 
The ISO 26262 associates safety requirements to recommendations for each risk level. These recommendations are described as “highly recommended” (++), “recommended” (+) and “not recommended” (o), where “highly recommended” indicates a need for implementation of the associated requirement for an application associated with the corresponding risk level. 
Since the ASIL is a well-defined classification scheme, we follow this risk level based categorization approach to classify each of the requirements according to their risk level definition. 
This allows for a risk based selection of a set of requirements for each individual mobility application.

During a homologation process, the integration of vehicle components is evaluated at each integration step. 
Therefore, the functional safety and security of the entire mobility system must be addressed during the requirements elicitation. 
Accordingly, we categorize the requirements whether they apply to the entire mobility system or the AI subsystem.

\subsection{Entire System Requirements}
\label{sec:requirements:system}

To provide more insight into the requirements elicitation process, we show an example requirement catered towards the entire system and its ASIL classification. 
REQ. 7 from \autoref{tab:requirements} ensures that the performance of the entire system is not affected under worst-case conditions. 
This can either encompass natural phenomena such as weather or lighting conditions, but also security threats for example by adversarial attacks or side-channel attacks. 
It is derived from the ASIL recommendation which states that the system shall be tested against worst-case errors. 
An example definition of a worst-case error is provided in \autoref{sec:implementation:application}.

\begin{table*}[tp]
	\caption{Three exemplary generic requirements and their ASIL risk level classification.}
	\label{tab:requirements} 
	\centering
	\begin{tabular}{c > {\centering} p{7.5cm} cccc}
		\toprule
		Identifier & Requirement & ASIL A & ASIL B &ASIL C & ASIL D \\
		\midrule
		Req. 7 	& The performance shall be compliant to the allowed worst-case error.	& ++ & ++ & ++ & ++ \\
		Req. 30	& The training, test and evaluation datasets shall be independent from each other. & ++ & ++ & ++ & ++ \\
		Req. 33	& The model's decisions shall be explained to aid the comparison between the modelling of the system and the trained model. & ++ & ++ & ++ & ++ \\
		\bottomrule	
	\end{tabular}
\end{table*}

\subsection{AI Subsystem Requirements}
\label{sec:requirements:AI}

REQ. 33 from \autoref{tab:requirements} is targeted towards the AI subsystem within the mobility system. 
This requirement is derived from an ASIL recommendation that states that the modelling of the system shall be compared to the resulting system. 
It is modified to fit the AI subsystem by stating that the model’s decisions shall be explained. 
This is because AI models are similar to black-boxes and there is low to no insight into how the model’s decisions are made. 
Therefore, specific explainability methods shall be used to gain insight into the correct functionality.

Analogously, REQ.30 states that the training, evaluation and testing datasets used during the development shall be independent from each other. 
This ensures that performance or training issues can be detected during the training phase.
The testing procedure for this requirement depends on the size and format of the datasets at hand.
An example evaluation of this requirement is explained in \autoref{sec:implementation:application}.

\subsection{Testability and Applicability}

The above-formulated requirements must be specified for each individual use case. 
To determine the effort needed to transfer a requirement between different mobility use cases we perform an analysis to determine the applicability and testability of each requirement. 
Additionally, we also provide indicators on the concretization effort between use cases and the type of test procedure. 
\autoref{tab:applicability} presents the categorization for the 3 example requirements from \autoref{sec:requirements:system} and \autoref{sec:requirements:AI}.

\begin{table*}[bp]
	\caption{Detailed analysis of the exemplary generic requirements from \autoref{tab:requirements}.}
	\label{tab:applicability} 
	\centering
	\begin{tabular}{ccccc}
		\toprule
		Identifier & Applicability & Concretization & Testability & Test Procedure \\
		\midrule
		Req. 7 & Complex & Major & High & Metric-based \\
		Req. 30 & Simple & Minor & High & Evidence-based \\
		Req. 33 & Complex & Minor & Medium & Metric-based \& Evidence-based \\
		\bottomrule	
	\end{tabular}
\end{table*}

\section{\uppercase{AI-based Systems in Mobility Applications}}
\label{sec:use-cases}

To assess the suitability of the generic audit requirements proposed in \autoref{sec:requirements}, we aim to perform practical tests for a concrete AI-based system.
To achieve universally applicable results, this system should be representative for different use cases in mobility applications and at the same time enable efficient initial tests. 
Thus, to select the most suitable exemplary use case we first introduce categories, which help to assess the suitability of different use cases. 
Following, we present a summary of possible AI-based use cases in mobility applications. 
Based on the introduced categories we then analyze all collected use cases to decide which use case is best suited for the initial practical tests of the audit requirements.

\subsection{Analysis Categories}
\label{sec:use-cases:categories}

To assess the suitability of different AI-based use cases in mobility applications for the practical audit requirements tests we choose five categories. 
These cover important aspects regarding the feasibility of the tests and the meaningfulness of the achieved results. 
In \autoref{tab:use-cases} these are later applied to individual use cases in mobility applications.

First, the relevance of each use case for the safety of the entire mobility system is rated as high, medium, low or none.
To achieve results for the auditing of the most critical use cases it is preferable to select a use case where a certain amount of safety relevance is given and a high relevance is ideal. 
This allows to develop audit criteria for critical tasks where an audit is important and required.

Next, the complexity and auditability of each use case is categorized as complex, medium or simple. 
This category covers the test effort required to derive the residual risk of an AI-based system implementing a use case.
For the initial practical tests, it is preferable to select a use case with rather low complexity.
This enables the most feasible development and to perform more extensive tests.

Third, the applicability of potential (adversarial) attacks on the AI component of each use case is rated as unrealistic, complex, medium or simple.
The most important factors that influence the categorization are the scalability of an attack, the availability of literature or demonstrations of an attack and the required access interface of an adversary.
Here, it is important that a direct attack interface to the AI component exists and it is best when attacks are comparatively simple to execute in practice.

Additionally, the required resources for implementing an exemplary system of each use case are rated as high, medium or low. 
Multiple factors like the availability of open-source datasets or representative implementations, the model size of involved AI components and the required computational resources for training or inference are considered. 
It is most suitable when datasets and implementations are publicly available and only low resources are required.

Lastly, the generalizability of the results of the practical audit requirements tests to other use cases is categorized as high, medium or low. 
Most importantly, this categorization considers which sensors are used, which perception components are involved and whether the use case impacts the planning or control of a vehicle. 
Use cases are preferable when some of these characteristics are shared with other use cases.

\subsection{Potential Use Cases}

Selecting a representative use case from the sheer number of potential use cases in AI-based mobility applications \cite{Yurtsever20, Ziebinski16} is a difficult task.
We tackle this by first collecting a list of ten high-level use cases which commonly occur in the specific areas of AD/ADAS.
For such use cases the need for audit requirements and procedures is highest, as associated systems are safety-relevant and are increasingly tested on public roads.

Concretely, we start by considering AD/ADAS use cases which have a direct impact on the control of a vehicle.
For all use cases we include the required perception of the respective road users or objects, meaning we do not only consider the final control algorithms.
Here, the first use case is termed \textit{collision avoidance}, which includes all functionalities that react to potential obstacles in the driving path of a vehicle, by initiating deceleration and/or steering motions. 
Notably, collision avoidance also contains (automatic) emergency breaking. 
Next, we consider the \textit{lane keeping} use case, which includes all functionalities that keep a vehicle in the current driving lane. 
Here, mainly steering motions are performed to tackle the given task. 
Further, we consider the \textit{lane changing} use case, which includes all functionalities that lead to a change in the driving lane of the vehicle.
Like lane keeping, steering motions are most important but also acceleration or deceleration motions are required to be able to merge in between two vehicles. 
Fourth, the\textit{ adaptive cruise control} use case includes functionalities that manage the distance to a vehicle driving in front of the ego vehicle.
Here, deceleration and acceleration motions are important to control the distance to the leading vehicle adaptively based on its driving maneuvers and speed.

After presenting use cases that directly affect the control of a vehicle, we now discuss additional use cases which have no direct control impact but are important to obtain a list of the most important AI-based AD/ADAS use cases. 
Therefore, the fifth use case is \textit{global path planning}. 
This includes functionalities to plan the global route/path of a vehicle, which consists of the rough path from the starting location to the target location. 
This route is updated online during driving depending on the current occupancy of roads or the probability of traffic jams. 
Next, the \textit{traffic sign assistant} use case includes all functionalities that show currently relevant traffic signs to the driver. 
However, this purely acts as an assistance feature and for example does not adapt the speed of a vehicle to the detected speed limit automatically. 
Additionally, we consider the\textit{ driver monitoring} use case.
Here, the goal is to detect drowsiness or distraction of a driver and provide a warning to the driver. 
This can reduce the number and criticality of accidents and is important when the driver must monitor assistance functionalities and be able to intervene rapidly.

In addition to the use cases that describe a specific functionality, we also consider basic use cases which provide information for various AD/ADAS-related functionalities. 
Here, the \textit{map-based localization} use case includes functionalities to determine the current position of a vehicle as it navigates through the environment. 
Typically, a map of the environment is used which can either be created dynamically or is created a priori.
Next, the \textit{road user detection} use case is considered, which includes functionalities to detect dynamic traffic participants like pedestrians, vehicles or cyclists. 
Lastly, the \textit{behavior prediction} use case includes functionalities to identify the behavior and subsequently the trajectory of traffic participants.
All three use cases have no direct impact on the control of a vehicle since they only provide information to functionalities for specialized use cases like collision avoidance or lane changing.

\begin{table*}[htb]
	\caption{Overview of considered AI-based use cases in mobility applications and their suitability for the practical testing of audit requirements. For each parameter, the symbol in brackets indicates whether this parameter value is suitable (↑), partially suitable (o) or unsuitable (↓) for the initial testing of the proposed audit requirements.}
	\label{tab:use-cases} 
	\centering
	\begin{tabular}{cccccc}
		\toprule
		Use Case 	& Safety 	& Complexity/	& Attack 		& Required 	& Generalizability \\
		& Relevance & Auditability 	& Applicability & Resources &  \\
		\midrule
		Collision Avoidance 	& High (↑) 		& Complex(o) & Medium (o) 		& High (↓) 		& High (↑) \\
		Lane Keeping 			& High (↑) 		& Medium (o) & Simple (↑)		& Medium (o)	& Medium (o) \\
		Lane Changing 			& High (↑) 		& Complex(o) & Medium (o) 		& High (↓) 		& High (↑) \\
		Adaptive Cruise Control & High (↑) 		& Medium (o) & Complex(o) 		& High (↓) 		& Medium (o) \\
		Global Path Planning 	& None (↓) 		& Simple (↑) & Unrealistic (↓) 	& High (↓)	 	& Low (o) \\
		Traffic Sign Assistant 	& Low (o)  		& Simple (↑) & Simple (↑) 		& Low (↑) 		& Medium (o) \\
		Driver Monitoring 		& Medium (o)	& Medium (o) & Unrealistic (↓) 	& Medium (o) 	& Low (o) \\
		Map-based Localization 	& High (↑)		& Medium (o) & Complex(o) 		& High (↓) 		& Low (o) \\
		Road User Detection 	& High (↑) 		& Complex(o) & Medium (o) 		& Medium (o) 	& Medium (o) \\
		Behavior Prediction 	& High (↑) 		& Complex(o) & Unrealistic (↓)  & Medium (o) 	& Low (o) \\
		\bottomrule	
	\end{tabular}
\end{table*}

\subsection{Use Case Selection}

After collecting a list of use cases in AI-based mobility applications we use the categories from \autoref{sec:use-cases:categories} to assess the suitability of each presented use case in \autoref{tab:use-cases}.
It is important to note that the assigned values in each category must be seen relative to each other.
For example, the value low only indicates that the use case is on the lower end when compared to all other presented use cases.

For the final selection of a use case, we start by dropping all use cases which do dot not fulfill the basic prerequisite for any category. 
This means that \textit{global path planning} is no longer considered since it has no direct safety relevance and it is unrealistic to apply attacks which target the AI component directly. 
Similarly, \textit{driver monitoring} and \textit{behavior prediction} are also no longer considered as it is very challenging or unrealistic for an adversary to apply attacks. 
Next, the use cases \textit{collision avoidance}, \textit{lane changing}, \textit{adaptive cruise control} and \textit{map-based localization} are considered as unsuitable because they typically require larger model sizes and less use case specific datasets are available for training and testing.

After dropping the seven unsuitable use cases only the three use cases \textit{lane keeping}, \textit{traffic sign assistant} and \textit{road user detection} are considered for further analysis. 
All three are in principle suitable and fulfill the basic prerequisites for all categories from \autoref{sec:use-cases:categories}. 
The remaining use cases differ with respect to their safety relevance (higher for \textit{lane keeping} and \textit{road user detection}) and their resource requirements/complexity (lower for \textit{traffic sign assistant}). 
To be able to test more audit requirements with the available resources we value the feasibility, i.e. the lower complexity, higher and select the \textit{traffic sign assistant} use case for further practical investigations. 
More complex and safety-relevant use cases can be explored later, once it is shown that the audit requirements can be applied and provide useful results. 
Details on the implementation of a system representing the selected use case are given later in \autoref{sec:implementation:setup}.

\section{\uppercase{Practical Implementation}}
\label{sec:implementation}

To test the applicability and expressiveness for real applications we implement the selected generic audit requirements from \autoref{sec:requirements} for the traffic sign assistant use case. 
Thus, in the following we first introduce the detailed architecture of the exemplary ADAS system which represents the traffic sign assistant functionality.
Then, we discuss the application of some interesting audit requirements and describe results and challenges.

\subsection{Experimental Setup}
\label{sec:implementation:setup}

For the traffic sign assistant use case, a single outside facing forward RGB camera sensor is used.
Based on the data of this sensor, the classification (and preceding detection) of traffic signs is performed using a DNN. 
This mimics the approach used for real traffic sign assistants \cite{Kwangyong17}. 
For our initial practical experiments, we only consider a DNN that performs a pure classification of traffic signs. 
The reason is that typically there exists a common detector for all kinds of road users and elements, like traffic signs, vehicles, pedestrian, etc. Based on the detected objects, the content of the detected bounding boxes is fed to special classification modules that specialize in concretely classifying the object in a box. 
For the case of a traffic sign assistant this means that a preceding road elements detector exists which outputs bounding boxes around detected signs. 
Then, a classifier that focusses on traffic signs is used to determine the exact sign type based on the given subset of the entire image selected by the preceding detector. 
The output of this system is therefore the detected traffic sign in the given image. 
In this work we use the classifier to test the proposed audit requirements in practice.

To train the DNN that classifies traffic signs we use the German Traffic Sign Recognition Benchmark (GTSRB) \cite{Stallkamp11} as training dataset. 
This dataset features 43 classes of German traffic signs. Using this dataset, we train a ResNet-18  \cite{He16} to represent the traffic sign assistant. 
ResNet-18 is selected because this architecture is one of the most successful architectures in DNN history and is often used in literature as a sensible baseline independent of the concrete task and use case. 
The training is performed without any augmentations and using only the standard GTSRB training dataset.

\subsection{Generic Toolbox}

To allow for an easy expansion of our tests to further mobility use cases, we design a toolbox in a modular way. 
In this work the toolbox is implemented in an exemplary fashion for the traffic sign assistant use case and some audit requirements discussed in \autoref{sec:implementation:application}. 
The goal is to continuously expand this toolbox\footnote{An overview of the toolbox is available at \url{www.bsi.bund.de/dok/1079914}.} and incorporate more audit requirements and use cases over time. 

\subsection{Application of Requirements}
\label{sec:implementation:application}

The generic catalogue of requirements that is elicited in \autoref{sec:requirements} enables a simple selection of requirements based on the risk level of the specific use case. 
We assume the traffic sign assistant to be an assistance system which is not able to gain automated control of the vehicle on its own.
However, as input to an AD system that for example regulates the vehicle’s speed in a certain range based on the detected traffic signs and additional parameters, the use case might be classified as ASIL A during homologation.
Due to this assumption, we select and specify requirements that are “highly recommended” (++) for the ASIL A risk level from our requirement catalogue \cite{P538}.

After the requirements are selected, the evaluation process consists of the following three steps for each requirement:
\begin{enumerate}
	\item Parameter selection: If the requirement requires the specification of parameters, the parameters are chosen according to the use case (if necessary by domain experts). Moreover, a rationale/justification how these parameters are derived is provided. 
	\item Description of the audit procedure: The audit procedure for the requirement is described. For “metric-based” requirements the technical evaluation/tests that are performed shall be described. For “evidence-based” requirements the procedural evaluation of evidence is described. 
	\item Verdict: The test results of “metric-based” tests or findings of “evidence-based” evaluations are assessed and a “pass” or “fail” verdict is given describing whether the requirement is fulfilled.
\end{enumerate}

In the following, we schematically show these steps for the exemplary requirements introduced in \autoref{sec:requirements} with the traffic sign assistant use case described in \autoref{sec:implementation:setup}.

\subsubsection{Requirement 7}
\begin{center}
	\textit{The performance shall be compliant to the allowed worst-case error.}
\end{center}

To fulfill this requirement the “performance” and “allowed worst-case error” have to be specified. 
In the case of the traffic sign assistant use case, we choose to measure the “performance” through the accuracy of the system. 
The “allowed worst-case error” is chosen as an accuracy greater than \SI{90}{\percent} and we take heavy rain as an example of a worst-case error. 
In the scope of this work, we schematically assume the vehicle running the traffic sign assistant is operated in Germany, where heavy rain is common and a \SI{90}{\percent} accuracy still offers sufficient reliability of the assistance system. 
This selection for the two parameters only serves as an example to demonstrate how the requirement can be tested in practice.
Depending on the boundary conditions, operational design domain, mitigation strategies or level of automation of the overall system it could be necessary to adapt the value for the required accuracy.
The rationale for the selection of these two parameters in a real-world use case requires a rationale from domain experts. 

The audit procedure for this requirement is “metric-based”, where a dataset (that the model was not trained on) containing data samples of all classes in heavy rain conditions is evaluated by the model. 
It is possible to use data samples captured in heavy rain conditions or transform data samples from clear weather conditions with a heavy rain simulation.
If the accuracy of this evaluation is greater than \SI{90}{\percent}, the requirement is fulfilled.
 
\begin{figure}[tb]
	\centering
	\begin{subfigure}{.49\linewidth}
		\centering
		\includegraphics[scale=0.6]{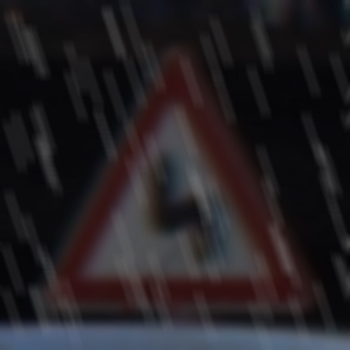}
		\caption{GTSRB class \num{21}}
		\label{fig:rain:21}
	\end{subfigure}
	\begin{subfigure}{.49\linewidth}
		\centering
		\includegraphics[scale=0.6]{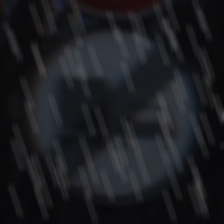}
		\caption{GTSRB class \num{42}}
		\label{fig:rain:41}
	\end{subfigure}
	\caption{Examples of the heavy rain transformation.}
	\label{fig:rain}
\end{figure}

Our toolbox implements a heavy rain transformation using the \textit{albumentations} library \cite{Buslaev20}. 
This allows to test the worst-case error on any suitable traffic sign dataset under heavy rain conditions.
We transform images from GTSRB using their heavy rain transformation, which for example results in images depicted in \autoref{fig:rain}. 
On \num{2580} data samples the system reaches an accuracy of $\sim$ \SI{79}{\percent}. 
Since the accuracy from the evaluation under heavy rain transformation is \SI{79}{\percent}, which is not greater than \SI{90}{\percent}, the requirement is not fulfilled and fails this evaluation.
For real-world use cases an assessment of domain experts is required regarding the representativity of different parameters of the used heavy rain transformation like the strength or structure of the rain drops.

Alternatively, a worst case could be represented by an adversarial attack to the system.
As an example we take a PGD attack \cite{Madry18} with a perturbation budget of \num{0.3}.
We repeat the outlined audit process but execute a PGD attack instead of applying a heavy rain transformation. 
Against this attack the system reaches an accuracy of $\sim$ \SI{21}{\percent} which means REQ. 7 is also not fulfilled using this second specification.
Note that PGD represents an attack in the digital domain.
In reality physical attacks which are applied in the environment itself pose a larger threat and testing against such attacks is more important.

\subsubsection{Requirement 30}
\begin{center}
\textit{The training, test and evaluation datasets shall be independent from each other.}
\end{center}

Since REQ. 30 has no parameters to be set, this step is skipped. 
The testing procedure of this requirement is classified as “evidence-based”. 
Hence, the dataset documentation, code and contents of each of the datasets shall be consulted. 
The documentation and code of the training procedure gives insight on how these datasets are generated. 
In this example, the evidence showed that the datasets were split before training the model into three disjoint datasets. 
Also, the datasets follow the same underlying distribution and are independent.
Therefore, the requirement is fulfilled.
It is important to use independent splits of the data to get a fair assessment of the quality of a model.
For example, images from a video recording of a single scene should not be used in different splits. 
Instead, images from a different recording like a different scene or in different weather must be used.

\subsubsection{Requirement 33}
\begin{center}
	\textit{The model’s decisions shall be explained to aid the comparison between the modelling of the system and the trained model.}
\end{center}

\begin{figure}[tb]
	\centering
	\begin{subfigure}{.49\linewidth}
		\centering
		\includegraphics[scale=0.6]{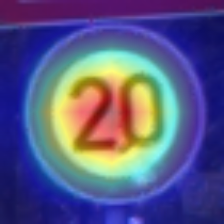}
		\caption{GTSRB class \num{0}}
		\label{fig:XAI:21}
	\end{subfigure}
	\begin{subfigure}{.49\linewidth}
		\centering
		\includegraphics[scale=0.6]{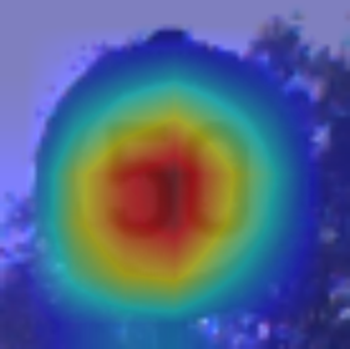}
		\caption{GTSRB class \num{2}}
		\label{fig:XAI:41}
	\end{subfigure}
	\caption{Examples of the GradCam explanation. The information with the highest influence on the model’s decision is highlighted in red.}
	\label{fig:XAI}
\end{figure}

For this requirement, the method used for explaining the decision and the system modelling the decisions have to be determined. 
In the schematic traffic sign assistant use case, we choose the following exemplary functional system requirement: \textit{The model decision on a traffic sign image shall depend on the figure displayed by the traffic sign, the signs coloration and/or the shape of the sign.}
In real-world applications depending on the chosen modelling, it would also be possible to implement automatic testing detecting whether a certain amount of background information is considered for the model’s decisions. 
In our case, we choose the \textit{GradCam} explainability method \cite{Selvaraju17} to explain a random set of \num{60} images per GTSRB class. 
\autoref{fig:XAI} presents some examples of a GradCam explanation on some images of the GTSRB dataset.
It clearly shows that the most important information for the decision (highlighted in red) of the model is based on the center of the image. 
We analyze all \num{60} images for each class and they show similar results. 
Hence, this evaluation is passed and the requirement is fulfilled.

\section{\uppercase{Conclusion}}
\label{sec:conclusion}

\subsection{Summary}

We introduce a list of generic audit requirements, which are technically relevant to assure the trustworthiness, security, safety, robustness, explainability, etc. of AI-based systems in mobility applications.
These requirements evolved under attention to existing regulations, norms, guidelines and an extensive literature review.
Additionally, we implement tools for exemplary audit requirements to demonstrate the applicability using a selected mobility application.
For this, we perform a comparison of different AD/ADAS use cases, based on various categories like complexity, auditability, available resources.
Using this analysis, we determined the traffic sign assistant use case to be best suited for the initial practical testing of the audit requirements.
Thus, we examine two exemplary DNNs trained on German traffic signs using the implemented audit requirements.
We find that the generic audit requirements can be specified to provide meaningful results on the DNN-based traffic sign assistants for different AI-specific properties.

\subsection{Outlook}

As discussed in \autoref{sec:implementation:application} we only use a subset of all proposed audit requirements for the initial practical tests.
A natural next step is to extend the practical tests to include all proposed requirements.
Additionally, one can expand the extent of the already implemented requirements.
Some of these requirements are quite extensive and can be implemented for practical tests in different ways.
In a follow-up work the exemplary implementation can be expanded to cover further aspects of the associated audit requirements.
This enables more extensive audits and increases the meaningfulness of the obtained results. 

Furthermore, it is especially interesting to test some audit requirements using actual hardware and test facilities. 
Instead of performing all tests in a simulation environment, the most interesting audit requirements should also be tested in reality.
Only these tests enable to properly assess the feasibility and expressiveness of the proposed audit requirements.

Additionally, the complexity of the audited system should be increased.
Instead of using only a DNN-based classifier, the system should be extended to be more representative of systems used in reality. 
Ideally, this is complemented by the application of the audit requirements to industry systems operating in practice.
This allows judging the applicability under real-world conditions and limitations.

Our goal is to continue this work and to consider at least one additional use case in addition to the traffic sign assistant.
We are working actively on the outlined next steps to further increase the meaningfulness of our results and refine the proposed requirements and best practices based on practical insights and limitations.	
We want to move towards applying the audit requirements in practice and create a formal technical guideline.
The obtained results could then be used as a blueprint for standardization activities and should be introduced to the relevant committees.

\section*{\uppercase{Acknowledgements}}

This work was supported by the Federal Office for Information Security (BSI), Germany, in project P538 \textit{AIMobilityAuditPrep}.

\bibliographystyle{apalike}
{\small
\bibliography{example}}

\begin{thebibliography}{}

\bibitem[AIMobilityAuditPrep, 2022]{P538}
AIMobilityAuditPrep (2022).
\newblock {AIMobilityAuditPrep: Final Results - Documentation}.
\newblock Technical report, German Federal Office for Information Security.
\newblock \url{www.bsi.bund.de/dok/1079912}.

\bibitem[ANSI/UL 4600, 2022]{ANSI4600}
ANSI/UL 4600 (2022).
\newblock {Standard for Safety for the Evaluation of Autonomous Products}.
\newblock Standard, American National Standards Institute, Underwriters
  Laboratories.

\bibitem[Berghoff et~al., 2020]{Berghoff20}
Berghoff, C., Neu, M., and von Twickel, A. (2020).
\newblock {Vulnerabilities of Connectionist AI Applications: Evaluation and
  Defence}.
\newblock {\em Frontiers in Big Data}, 3.

\bibitem[Buslaev et~al., 2020]{Buslaev20}
Buslaev, A., Parinov, A., Khvedchenya, E., Iglovikov, V., and Kalinin, A.
  (2020).
\newblock {Albumentations: Fast and Flexible Image Augmentations}.
\newblock {\em Information}, 11.

\bibitem[DIN Roadmap AI, 2022]{DINAI}
DIN Roadmap AI (2022).
\newblock {German Standardization Roadmap on Artificial Inteligence}.
\newblock Draft, German Institute for Standardization.

\bibitem[EU AI Act, 2021]{AIAct}
EU AI Act (2021).
\newblock {Regulation of the European Parliament and of the Council Laying Down
  Harmonised Rules on Artificial Intelligence and Amending Certain Union
  Legislative Acts}.
\newblock Draft, European Commission.

\bibitem[Geirhos et~al., 2020]{Geirhos20}
Geirhos, R., Jacobsen, J., Michaelis, C., Zemel, R., Brendel, W., Bethge, M.,
  and Wichmann, F.~A. (2020).
\newblock {Shortcut Learning in Deep Neural Networks}.
\newblock {\em Nature Machine Intelligence}, 2:665--673.

\bibitem[Gilpin et~al., 2018]{Gilpin18}
Gilpin, L., Bau, D., Yuan, B., and Bajwa, A. (2018).
\newblock {Explaining Explanations: An Overview of Interpretability of Machine
  Learning}.
\newblock In {\em International Conference on Data Science and Advanced
  Analytics}, Turin, Italy.

\bibitem[Goldblum et~al., 2020]{Goldblum20}
Goldblum, M., Tsipras, D., Xie, C., and Chen, X. (2020).
\newblock {Dataset Security for Machine Learning: Data Poisoning, Backdoor
  Attacks, and Defenses}.
\newblock {\em IEEE Transactions on Pattern Analysis \& Machine Intelligence}.

\bibitem[Guidotti et~al., 2018]{Guidotti18}
Guidotti, R., Monreale, A., Ruggieri, S., Turini, F., Giannotti, F., and
  Pedreschi, D. (2018).
\newblock {A Survey Of Methods for Explaining Black Box Models}.
\newblock {\em ACM Computing Surveys}, 51:1--42.

\bibitem[He et~al., 2016]{He16}
He, K., Zhang, X., Ren, S., and Sun, J. (2016).
\newblock {Deep Residual Learning for Image Recognition}.
\newblock In {\em Conference on Computer Vision and Pattern Recognition}, Las
  Vegas, USA.

\bibitem[Hendrycks and Dietterich, 2019]{Hendrycks19}
Hendrycks, D. and Dietterich, T. (2019).
\newblock {Benchmarking Neural Network Robustness to Common Corruptions and
  Perturbations}.
\newblock In {\em International Conference on Learning Representations}, New
  Orleans, USA.

\bibitem[ISO 21448, 2022]{ISO21448}
ISO 21448 (2022).
\newblock {Road vehicles - Safety of the intended funtionality}.
\newblock Standard, International Organization for Standardization.

\bibitem[ISO 26262, 2018]{ISO26262}
ISO 26262 (2018).
\newblock {Road vehicles - Functional safety}.
\newblock Standard, International Organization for Standardization.

\bibitem[ISO/AWI PAS 8800, 2022]{ISO8800}
ISO/AWI PAS 8800 (2022).
\newblock {Road vehicles - Safety and artificial intelligence}.
\newblock Standard, International Organization for Standardization.

\bibitem[ISO/AWI TS 5083, 2022]{ISO5083}
ISO/AWI TS 5083 (2022).
\newblock {Road vehicles - Safety for automated driving systems - Design,
  verification and validation}.
\newblock Standard, International Organization for Standardization.

\bibitem[ISO/IEC DTR 5469, 2022]{ISO5469}
ISO/IEC DTR 5469 (2022).
\newblock {Artificial intelligence - Functional safety and AI systems}.
\newblock Standard, International Organization for Standardization,
  International Electrotechnical Commission.

\bibitem[ISO/IEC PRF TS 4213, 2022]{ISO4213}
ISO/IEC PRF TS 4213 (2022).
\newblock {Artificial intelligence - Assessment of machine learning
  classification performance}.
\newblock Standard, International Organization for Standardization,
  International Electrotechnical Commission.

\bibitem[ISO/IEC TR 24028, 2020]{ISO24028}
ISO/IEC TR 24028 (2020).
\newblock {Artificial intelligence - Overview of trustworthiness in artificial
  intelligence}.
\newblock Standard, International Organization for Standardization,
  International Electrotechnical Commission.

\bibitem[ISO/IEC TR 24029-1, 2021]{ISO24029}
ISO/IEC TR 24029-1 (2021).
\newblock {Artificial intelligence - Assessment of the robustness of neural
  networks - Part 1: Overview}.
\newblock Standard, International Organization for Standardization,
  International Electrotechnical Commission.

\bibitem[ISO/SAE 21434, 2021]{ISO21434}
ISO/SAE 21434 (2021).
\newblock {Road vehicles - Cybersecurity engineering}.
\newblock Standard, International Organization for Standardization, SAE
  International.

\bibitem[Karpathy, 2021]{tesla}
Karpathy, A. (2021).
\newblock {Workshop on Autonomous Driving - Tesla Keynote}.
\newblock In {\em Conference on Computer Vision and Pattern Recognition},
  Nashville, USA.

\bibitem[Li et~al., 2022]{Li22}
Li, B., Qi, P., Liu, B., Di, S., Liu, J., Pei, J., Yi, J., and Zhou, B. (2022).
\newblock {Trustworthy AI: From Principles to Practices}.
\newblock {\em ACM Computing Surveys}.

\bibitem[Lim et~al., 2017]{Kwangyong17}
Lim, K., Hong, Y., Choi, Y., and Byun, H. (2017).
\newblock {Real-time Traffic Sign Recognition based on a General Purpose GPU
  and Deep Learning}.
\newblock {\em PLOS ONE}, 12:1--22.

\bibitem[Madry et~al., 2018]{Madry18}
Madry, A., Makelov, A., Schmidt, L., Tsipras, D., and Vladu, A. (2018).
\newblock {Towards Deep Learning Models Resistant to Adversarial Attacks}.
\newblock In {\em International Conference on Learning Representations},
  Vancouver, Canada.

\bibitem[Mohseni et~al., 2020]{Mohseni20}
Mohseni, S., Pitale, M., Singh, V., and Wang, Z. (2020).
\newblock {Practical Solutions for Machine Learning Safety in Autonomous
  Vehicles}.
\newblock In {\em Conference on Artificial Intelligence: Workshop on Safe AI},
  New York, USA.

\bibitem[Orekondy et~al., 2019]{Orekondy19}
Orekondy, T., Schiele, B., and Fritz, M. (2019).
\newblock {Knockoff Nets: Stealing Functionality of Black-Box Models}.
\newblock In {\em Conference on Computer Vision and Pattern Recognition}, Long
  Beach, USA.

\bibitem[Papernot et~al., 2017]{Papernot17}
Papernot, N., McDaniel, P., Goodfellow, I., Jha, S., Celik, B., and Swami, A.
  (2017).
\newblock {Practical Black-Box Attacks against Machine Learning}.
\newblock In {\em ACM Asia Conference on Computer and Communications Security},
  Abu Dhabi, United Arab Emirates.

\bibitem[Radlak et~al., 2020]{Radlak20}
Radlak, K., Szczepankiewicz, M., Jones, T., and Serwa, P. (2020).
\newblock {Organization of Machine Learning based Product Development as per
  ISO 26262 and ISO/PAS 21448}.
\newblock In {\em Pacific Rim International Symposium on Dependable Computing},
  Perth, Australia.

\bibitem[Raji et~al., 2020]{Raji20}
Raji, I.~D., Smart, A., White, R., Mitchell, M., Gebru, T., Hutchinson, B.,
  Smith-Loud, J., Theron, D., and Barnes, P. (2020).
\newblock {Closing the AI Accountability Gap: Defining an End-to-End Framework
  for Internal Algorithmic Auditing}.
\newblock In {\em ACM Conference on Fairness, Accountability, and
  Transparency}, Barcelona, Spain.

\bibitem[SAE J3016, 2014]{SAEJ3016}
SAE J3016 (2014).
\newblock {Levels of Driving Automation}.
\newblock Standard, SAE International.

\bibitem[Schwarzschild et~al., 2021]{Schwarzschild21}
Schwarzschild, A., Goldblum, M., Gupta, A., Dickerson, J., and Goldstein, T.
  (2021).
\newblock {Just How Toxic is Data Poisoning? A Unified Benchmark for Backdoor
  and Data Poisoning Attacks}.
\newblock In {\em International Conference on Machine Learning}, Vienna,
  Austria.

\bibitem[Selvaraju et~al., 2017]{Selvaraju17}
Selvaraju, R., Cogswell, M., Das, A., Vedantam, R., Parikh, D., and Batra, D.
  (2017).
\newblock {Grad-CAM: Visual Explanations from Deep Networks via Gradient-based
  Localization}.
\newblock In {\em International Conference on Computer Vision}, Venice, Italy.

\bibitem[Stallkamp et~al., 2011]{Stallkamp11}
Stallkamp, J., Schlipsing, M., Salmen, J., and Igel, C. (2011).
\newblock {Man vs. Computer: Benchmarking Machine Learning Algorithms for
  Traffic Sign Recognition}.
\newblock In {\em International Joint Conference on Neural Networks}, San Jose,
  USA.

\bibitem[Szegedy et~al., 2014]{Szegedy14}
Szegedy, C., Zaremba, W., Sutskever, I., Bruna, J., Erhan, D., Goodfellow, I.,
  and Fergus, R. (2014).
\newblock {Intriguing Properties of Neural Networks}.
\newblock In {\em International Conference on Learning Representations}, Banff,
  Canada.

\bibitem[UNECE R 155, 2021]{UNECER155}
UNECE R 155 (2021).
\newblock {Uniform provisions concerning the approval of vehicles with regards
  to cyber security and cyber security management system}.
\newblock Standard, United Nations Economic Commission for Europe.

\bibitem[Waymo, 2021]{waymo}
Waymo (2021).
\newblock {How we’ve built the World’s Most Experienced Urban Driver}.
\newblock {\em Waypoint}.

\bibitem[Yurtsever et~al., 2020]{Yurtsever20}
Yurtsever, E., Lambert, J., Carballo, A., and Takeda, K. (2020).
\newblock {A Survey of Autonomous Driving: Common Practices and Emerging
  Technologies}.
\newblock {\em IEEE Access}, 8:58443--58469.

\bibitem[Ziebinski et~al., 2016]{Ziebinski16}
Ziebinski, A., Cupek, R., Erdogan, H., and Waechter, S. (2016).
\newblock {A Survey of ADAS Technologies for the Future Perspective of Sensor
  Fusion}.
\newblock In {\em International Conference on Computational Collective
  Intelligence}, Halkidiki, Greece.

\end{thebibliography}

\end{document}